%% file: main.tex
\documentclass[10pt,twocolumn,letterpaper]{article}

\usepackage[pagenumbers]{cvpr} % To force page numbers, e.g. for an arXiv version

\usepackage{graphicx}
\usepackage{amsmath}
\usepackage{amssymb}
\usepackage{booktabs}

\usepackage{caption}
\usepackage{adjustbox}
\usepackage{wrapfig}
\usepackage{multirow}
\usepackage{kotex}
\usepackage{comment}

\usepackage{algorithm}
\usepackage{algpseudocode}
\usepackage[numbers,sort]{natbib}
\usepackage{verbatim}
\usepackage{xcolor}
\usepackage[accsupp]{axessibility}

\newcommand\blfootnote[1]{%
  \begingroup
  \renewcommand\thefootnote{}\footnote{#1}%
  \addtocounter{footnote}{-1}%
  \endgroup
}

\usepackage[pagebackref,breaklinks,colorlinks]{hyperref}

\usepackage[capitalize]{cleveref}
\crefname{section}{Sec.}{Secs.}
\Crefname{section}{Section}{Sections}
\Crefname{table}{Table}{Tables}
\crefname{table}{Tab.}{Tabs.}

\def\cvprPaperID{00005} % *** Enter the CVPR Paper ID here
\def\confName{CVPR}
\def\confYear{2023}

\begin{document}

\title{Custom-Edit: Text-Guided Image Editing with Customized Diffusion Models}

\author{Jooyoung Choi$^1$ ~~~~~~~ Yunjey Choi$^2$ ~~~~~~~ Yunji Kim$^2$  ~~~~~~~ Junho Kim$^{2, *}$ ~~~~~~~ Sungroh Yoon$^{1, *}$\\
$^1$ Data Science and AI Laboratory, ECE, Seoul National University\\
$^2$ NAVER AI Lab\\
}
%\and
\maketitle
\blfootnote{$*$ Corresponding Authors}

%%%%%%%%% ABSTRACT
\begin{abstract}
   Text-to-image diffusion models can generate diverse, high-fidelity images based on user-provided text prompts. Recent research has extended these models to support text-guided image editing. While text guidance is an intuitive editing interface for users, it often fails to ensure the precise concept conveyed by users. To address this issue, we propose \textit{Custom-Edit}, in which we (i) customize a diffusion model with a few reference images and then (ii) perform text-guided editing. Our key discovery is that customizing only language-relevant parameters with augmented prompts improves reference similarity significantly while maintaining source similarity. Moreover, we provide our recipe for each customization and editing process. We compare popular customization methods and validate our findings on two editing methods using various datasets.
\end{abstract}

%%%%%%%%% BODY TEXT
\vspace{-0.8em}
\section{Introduction}
\label{sec:intro}
\begin{figure}[t]
\begin{center}
   \includegraphics[width=1.0\linewidth]{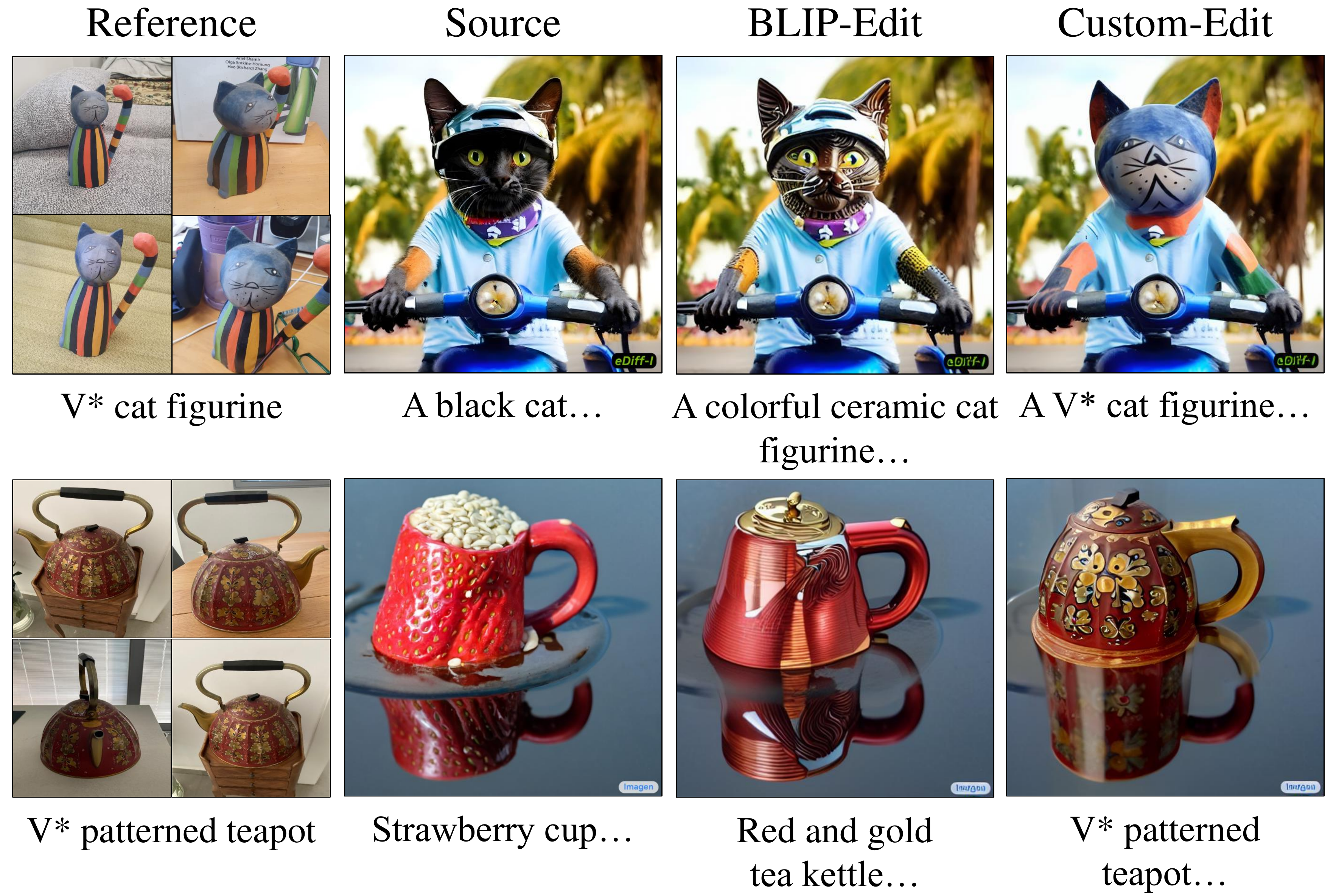}
\end{center}
\vspace*{-4mm}
   \caption{Our \textit{Custom-Edit} allows high-fidelity text-guided editing, given a few references. Edited images with BLIP2~\cite{li2023blip} captions show the limitation of textual guidance in capturing the fine-grained appearance of the reference.}
   \vspace{-1em}
\label{fig:blip}

\end{figure}

Recent work on deep generative models has led to rapid advancements in image editing. Text-to-image models~\cite{rombach2022high,sahariaphotorealistic} trained on large-scale databases~\cite{schuhmannlaion} allow intuitive editing~\cite{hertz2022prompt,meng2021sdedit} of images in various domains. Then, to what extent can these models support precise editing instructions? Can a unique concept of the user, especially one not encountered during large-scale training, be utilized for editing? Editing with a prompt acquired from a well-performing captioning model~\cite{li2023blip} fails to capture the appearance of reference, as shown in \cref{fig:blip}.

We propose \textit{Custom-Edit}, a two-step approach that involves (i) customizing the model~\cite{gal2022image,ruiz2022dreambooth,kumari2022multi} using a few reference images and then (ii) utilizing effective text-guided editing methods~\cite{hertz2022prompt,mokady2022null,meng2021sdedit} to edit images. While prior customization studies~\cite{gal2022image,ruiz2022dreambooth,kumari2022multi} deal with the random generation of images (noise$\rightarrow$image), our work focuses on image editing (image$\rightarrow$image). As demonstrated in \cref{fig:blip}, customization improves faithfulness to the reference's appearance by a large margin. This paper shows that customizing only language-relevant parameters with augmented prompts significantly enhances the quality of edited images. Moreover, we present our design choices for each customization and editing process and discuss the \textit{source-reference trade-off} in Custom-Edit.

\begin{figure*}
\centering
\includegraphics[width=1.0\linewidth]{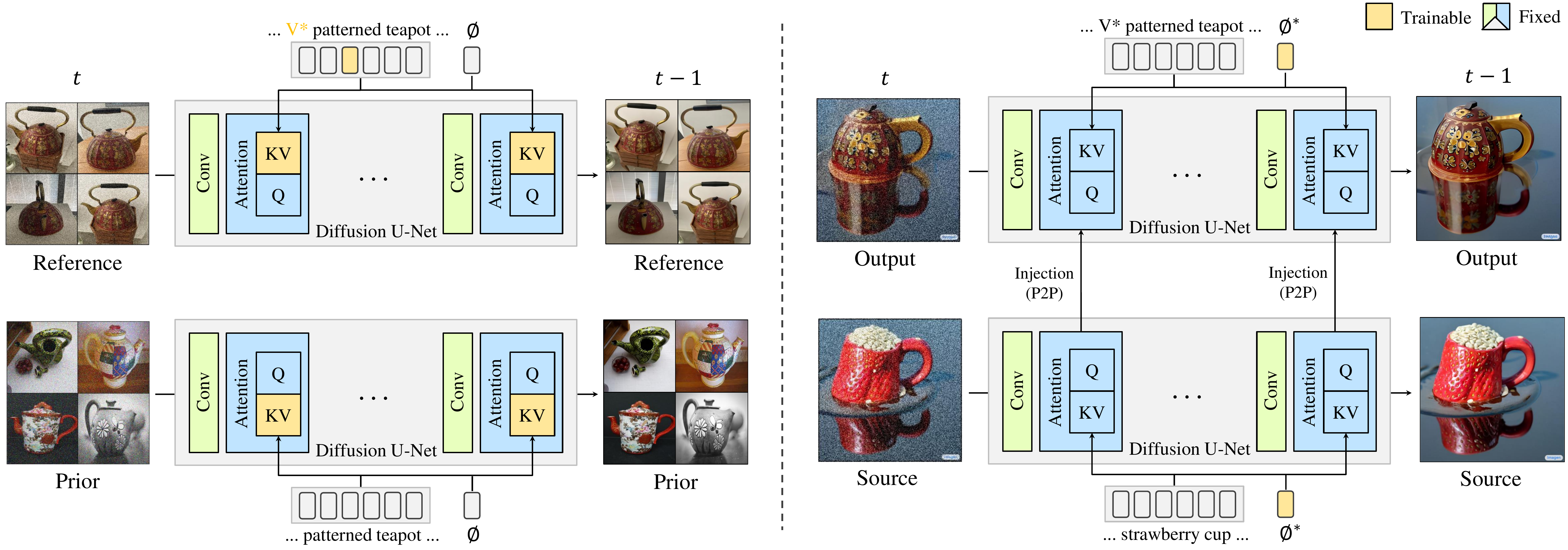} \\
\vspace{0mm}
\makebox[0.5\linewidth][c]{\footnotesize{(a) Customization process}}\hfill
\makebox[0.5\linewidth][c]{\footnotesize{(b) Editing process}}\hfill \\
\vspace{-2mm}
   \caption{Our Custom-Edit consists of two processes: the customization process and the editing process. \textbf{(a) Customization.} We customize a diffusion model by optimizing only language-relevant parameters (i.e., custom embedding V* and attention weights) on a given set of reference images. We also apply the prior preservation loss to alleviate the language drift. \textbf{(b) Editing.} We then transform the source image to the output using the customized word. We leverage the P2P and Null-text inversion methods~\cite{hertz2022prompt, mokady2022null} for this process.}
\label{fig:pipeline}
\vspace{-2mm}
\end{figure*}

\section{Diffusion Models}

Throughout the paper, we use Stable Diffusion~\cite{rombach2022high}, an open-source text-to-image model. The diffusion model~\cite{ho2020denoising,sohl2015deep,song2019generative,dhariwal2021diffusion} is trained in the latent space of a VAE~\cite{kingma2013auto}, which downsamples images for computation efficiency. 
The model is trained to reconstruct the clean latent representation $x_0$ from a perturbed representation $x_t$ given the text condition $c$, which is embedded with the CLIP text encoder~\cite{radford2021learning}. The diffusion model is trained with the following objective:
\vspace{-0.5em}
\begin{equation}\label{eq:loss}
\sum_{t=1}^{T} \mathbb{E}_{x_0,\epsilon}[||\epsilon-\epsilon_\theta(x_t, t, c)||^2],
\end{equation}
where $\epsilon$ is an added noise, $t$ is a time step indicating a perturbed noise level, and $\epsilon_{\theta}$ is a diffusion model with a U-Net~\cite{ronnebefrger2015u} architecture with attention blocks~\cite{vaswani2017attention}. 
During training, the text embeddings are projected to the keys and values of cross-attention layers, and the text encoder is kept frozen to preserve its \textit{language understanding capability}.
Imagen~\cite{sahariaphotorealistic} and eDiffi~\cite{balaji2022ediffi} have shown that leveraging rich language understandings of large language models by freezing them is the key to boosting the performance.

\section{Custom-Edit}
Our goal is to edit images with complex visual instructions given as reference images (\cref{fig:blip}). 
Therefore, we propose a two-step approach that (i) customizes the model on given references (\cref{sec:3.1}) and (ii) edits images with textual prompts (\cref{sec:3.2}). 
Our method is presented in \cref{fig:pipeline}.

\subsection{Customization}
\label{sec:3.1}
\noindent\textbf{Trainable Parameters.}
We optimize only the keys and values of cross-attention and the `[rare token]', following Custom-Diffusion~\cite{kumari2022multi}.
As we discuss in \cref{sec:experiment}, our results indicate that training these \textit{language-relevant} parameters is crucial for successfully transferring reference concepts to source images. Furthermore, training only these parameters requires less storage than Dreambooth~\cite{ruiz2022dreambooth}.

\noindent\textbf{Augmented Prompts.} We fine-tune the abovementioned parameters by minimizing \cref{eq:loss}. We improve Custom-Diffusion for editing by augmenting the text input as `[rare token] [\textit{modifier}] [class noun]' (e.g., `V* \textit{patterned} teapot'). We find that `[modifier]' encourages the model to focus on learning the appearance of the reference.

\noindent\textbf{Datasets.} 
To keep the language understanding while fine-tuning on the reference, we additionally minimize prior preservation loss~\cite{ruiz2022dreambooth} over diverse images belonging to the same class as the reference. Thus, we use  CLIP-retrieval~\cite{beaumont-2022-clip-retrieval} to retrieve 200 images and their captions from the LAION dataset~\cite{schuhmannlaion} using the text query `photo of a [modifier] [class noun]'.

\subsection{Text-Guided Image Editing}
\label{sec:3.2} 

\begin{figure*}
\begin{center}
   \includegraphics[width=0.85\linewidth]{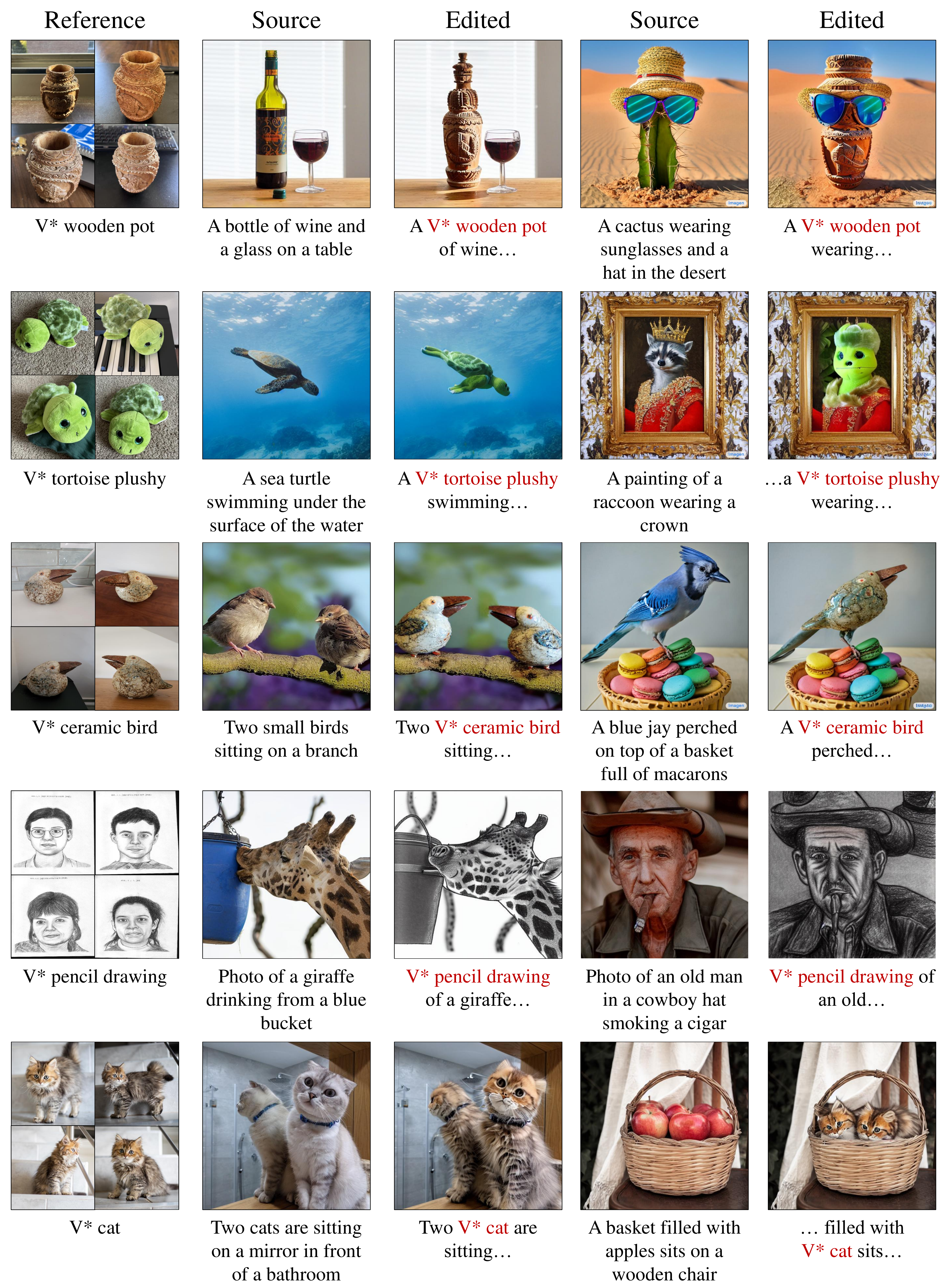}
\end{center}
   \caption{\textbf{Custom-Edit results.} Our method transfers the reference's appearance to the source image with unprecedented fidelity. The structures of the source are well preserved. We obtain source prompts using BLIP2~\cite{li2023blip}. Except for the pencil drawing example, we use local editing of P2P with automatically generated masks.}
\label{fig:main}
\end{figure*}

\noindent\textbf{Prompt-to-Prompt.}
We use Prompt-to-Prompt~\cite{hertz2022prompt} (P2P), a recently introduced editing framework that edits images by only modifying source prompts. P2P proposes attention injection to preserve the structure of a source image.
For each denoising step $t$, let us denote the attention maps of the source and edited image as $M_t$ and ${M_t}^*$, respectively. P2P then injects a new attention map $Edit(M_t,{M_t}^*,t)$ into the model $\epsilon_{\theta}$.
$Edit$ is an attention map editing operation, including \textit{prompt refinement} and \textit{word swap}.
Additionally, P2P enables local editing with an automatically computed mask. P2P computes the average of cross-attention $\bar{M}_{t,w}$ and ${\bar{M}_{t,w}}^*$ related to the word $w$ and thresholds them to produce the binary mask $B(\bar{M_t}) \cup B({\bar{M}_t}^*)$. Before editing with P2P, we utilize Null-Text Inversion~\cite{mokady2022null} to boost the source preservation. 
%Refer to Sec. \textcolor{red}{C} for a more description.
Refer to~\cref{sec:appendix-background} for a more description.

\noindent\textbf{Operation Choice.} 
Due to the limited number of reference images, the customized words favor only a limited variety of structures. This inspired us to propose the following recipe. First, we use \textit{prompt refinement} for the Edit function. \textit{Word swap} fails when the customized words do not prefer the swapped attention map. Second, we use mask $B(\bar{M_t})$ rather than $B(\bar{M_t}) \cup B({\bar{M}_t}^*)$, as the customized words are likely to generate incorrect masks.

\noindent\textbf{Source-Reference Trade-Off.} 
A key challenge in image editing is balancing the edited image's source and reference similarities. We refer to $\tau / T$ as \textit{strength}, where P2P injects self-attention from $t=T$ to $t=\tau$.
In P2P, we observed that 
a critical factor in controlling the trade-off is the injection of self-attention rather than cross-attention. Higher strength denotes higher source similarity at the expense of reference similarity.
In \cref{sec:experiment}, we also show results with SDEdit~\cite{meng2021sdedit}, which diffuses the image from $t=0$ to $t=\tau$ and denoises it back. As opposed to P2P, higher strength in SDEdit means higher reference similarity. 

\section{Experiment}
\label{sec:experiment}
In this section, we aim to validate each process of Custom-Edit. Specifically, we assess our design choices for customization by using Textual Inversion~\cite{gal2022image} and Dreambooth~\cite{ruiz2022dreambooth} in the customization process. 
We compare their source-reference trade-off in the editing process. As well as P2P, we use SDEdit~\cite{meng2021sdedit} for experiments.

\noindent\textbf{Baselines.} Textual Inversion learns a new text embedding V*, initialized with a class noun (e.g., `pot'), by minimizing \cref{eq:loss} for the input prompt `V*'.
Dreambooth fine-tunes the diffusion model while the text encoder is frozen. \cref{eq:loss} is minimized over a few images given for input prompt `[rare token] [class noun]' (e.g., `ktn teapot'). 
SDEdit is the simplest editing method that diffuse-and-denoise the image.

\noindent\textbf{Datasets.}
We use eight references in our experiments, including two pets, five objects, and one artwork. For each reference, we used five source images on average.

\noindent\textbf{Metrics.}
We measure the source and reference similarities with CLIP ViT-B/32~\cite{radford2021learning}.
We use strengths [0.2, 0.4, 0.6, 0.8] for P2P and [0.5, 0.6, 0.7, 0.8] for SDEdit results. 
We generated two P2P samples with cross-attention injection strengths [0.2, 0.6], and three SDEdit samples for each strength and source image from different random seeds.

\noindent\textbf{Inference Details.}
We employ a guidance scale of 7.5 and 50 inference steps. We acquire all source prompts using BLIP2~\cite{li2023blip}. 
%More details are available in Sec. \textcolor{red}{B}.
More details are available in~\cref{sec:appendix-details}.
 
\subsection{Qualitative Results}

\cref{fig:main} illustrates the selected results. Custom-Edit transfers the reference's detailed appearance to the source while preserving the overall structure. For example, Custom-Edit generates a horizontally elongated V* wooden pot from the wine bottle (first row). In the second row, Custom-Edit generates a V* tortoise plushy wearing a hat with the texture of its shell. The blue jay in the third row became a V* ceramic bird with perfectly preserved macarons. In the last row, the V* cat is sitting in a pose that does not exist in the reference set. 
%We show qualitative comparisons in Sec. \textcolor{red}{A.1}.
We show qualitative comparisons in~\cref{sec:appendix-qualitative}.

\begin{figure}[t]
\begin{center}
   \includegraphics[width=0.78\linewidth]{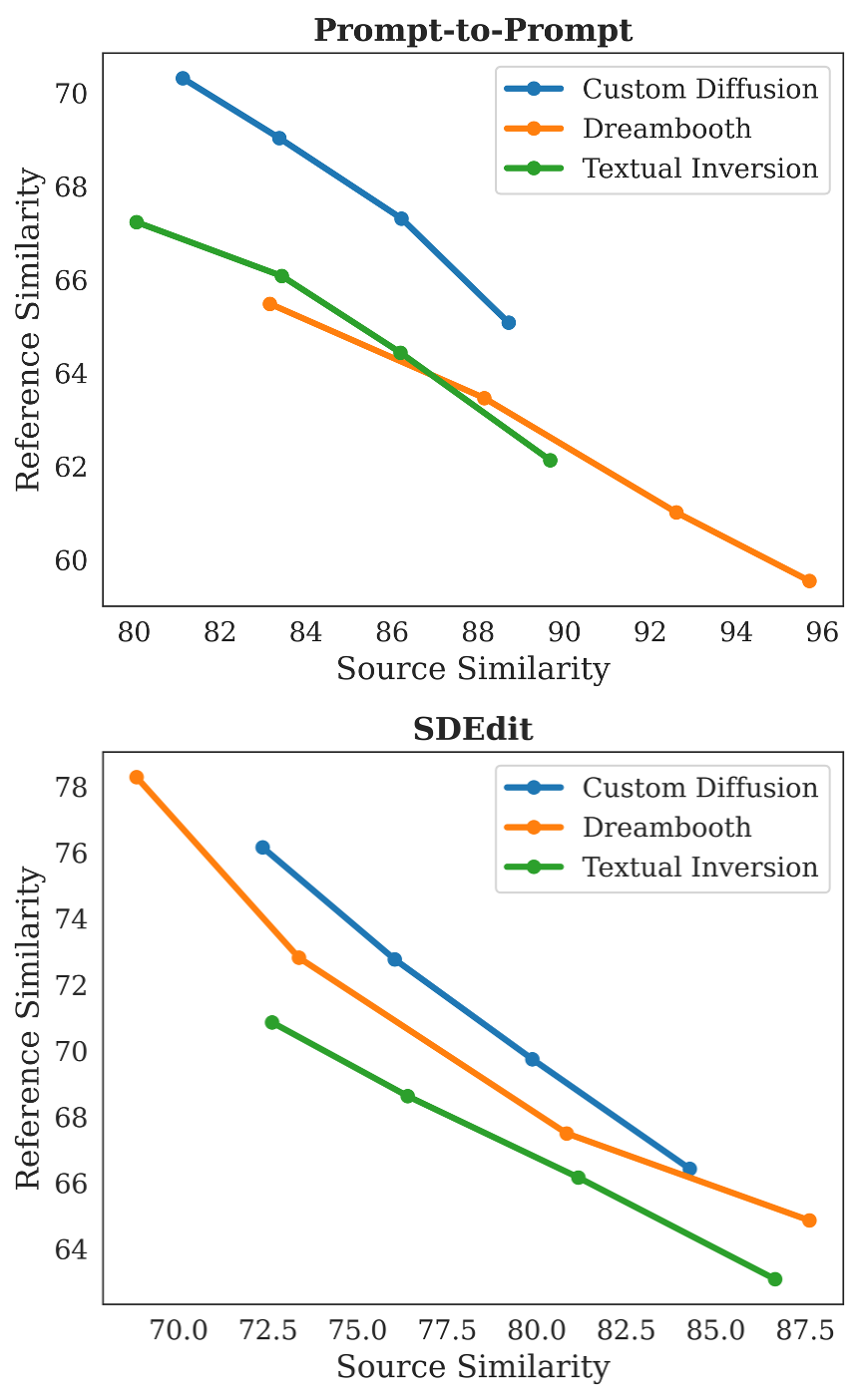}
\end{center}
\vspace{-4mm}
   \caption{\textbf{Source-Reference Trade-Off.} Custom-Diffusion shows the best trade-off, indicating the effectiveness of training only language-relevant parameters. 
   We exhibit qualitative comparisons and samples with various strengths in \cref{sec:appendix-strength}.
   }

\label{fig:ref-tradeoff}
\end{figure}

\subsection{Quantitative Results}

\cref{fig:ref-tradeoff} shows average trade-off curves on P2P and SDEdit. Our improved Custom-Diffusion yields the best trade-off, while Textual Inversion shows similar source similarity but lower reference similarity. Dreambooth has higher source similarity but lower reference similarity, suggesting that it is ineffective in modifying images. SDEdit results also show a similar tendency, supporting our claim that customizing language-relevant parameters is effective for editing. Note that SDEdit shows lower source similarity than P2P, indicating the superiority of P2P and our operation choices in text-guided editing.

\section{Discussion}
We propose Custom-Edit, which allows fine-grained editing with textual prompts. We present our design choices for each process, which can benefit future customization and editing work. Additionally, we discuss the trade-off between source and reference in diffusion-based editing.

Although Custom-Edit shows various successful editing results, there are some failure cases, as presented 
%in Sec. \textcolor{red}{A.3}.
in~\cref{sec:appendix-failure}. 
Custom-Edit sometimes edits undesired regions or fails to edit complex backgrounds. We hypothesize that this is due to the inaccurate attention maps of Stable Diffusion~\cite{hertz2022prompt,mokady2022null} and the limited controllability of the text input. Potential solutions are to apply Custom-Edit on text-to-image models with larger text encoders~\cite{sahariaphotorealistic,balaji2022ediffi} or extended controllability~\cite{li2023gligen,zhang2023adding}.

\vspace{1em}
\noindent \textbf{Acknowledgements: }
This work was supported by the National Research Foundation of Korea (NRF) grants funded by the Korea government (Ministry of Science and ICT, MSIT) (2022R1A3B1077720), Institute of Information \& communications Technology Planning \& Evaluation (IITP) grants funded by the Korea government (MSIT) (2021-0-01343: AI Graduate School Program, SNU), and the BK21 FOUR program of the Education and Research Program for Future ICT Pioneers, Seoul National University in 2023.

%%%%%%%%% REFERENCES
{\small
\bibliographystyle{ieee_fullname}
\bibliography{egbib}
}

%%%%%%%%% APPENDIX
\input{Appendix}

\end{document}

%% file: Appendix.tex
%\clearpage
%\newpage

%\twocolumn[
%]

% \twocolumn[
% \centering
% \Large
% \textbf{Custom-Edit: Text-Guided Image Editing with Customized Diffusion Models} \\
% \vspace{0.5em}-- Supplementary Material --\\
% \vspace{1.0em}
% ]

\setcounter{section}{0}
\renewcommand\thesection{\Alph{section}}
\setcounter{table}{0}
\renewcommand{\thetable}{\Alph{table}}
\setcounter{figure}{0}
\renewcommand{\thefigure}{\Alph{figure}}
\setcounter{equation}{0}
\renewcommand{\theequation}{\Alph{equation}}

\section{Additional Results}
\label{sec:appendix-additional}
\begin{figure*}
\begin{center}
%\fbox{\rule{0pt}{2in} \rule{.9\linewidth}{0pt}}
   \includegraphics[width=0.85\linewidth]{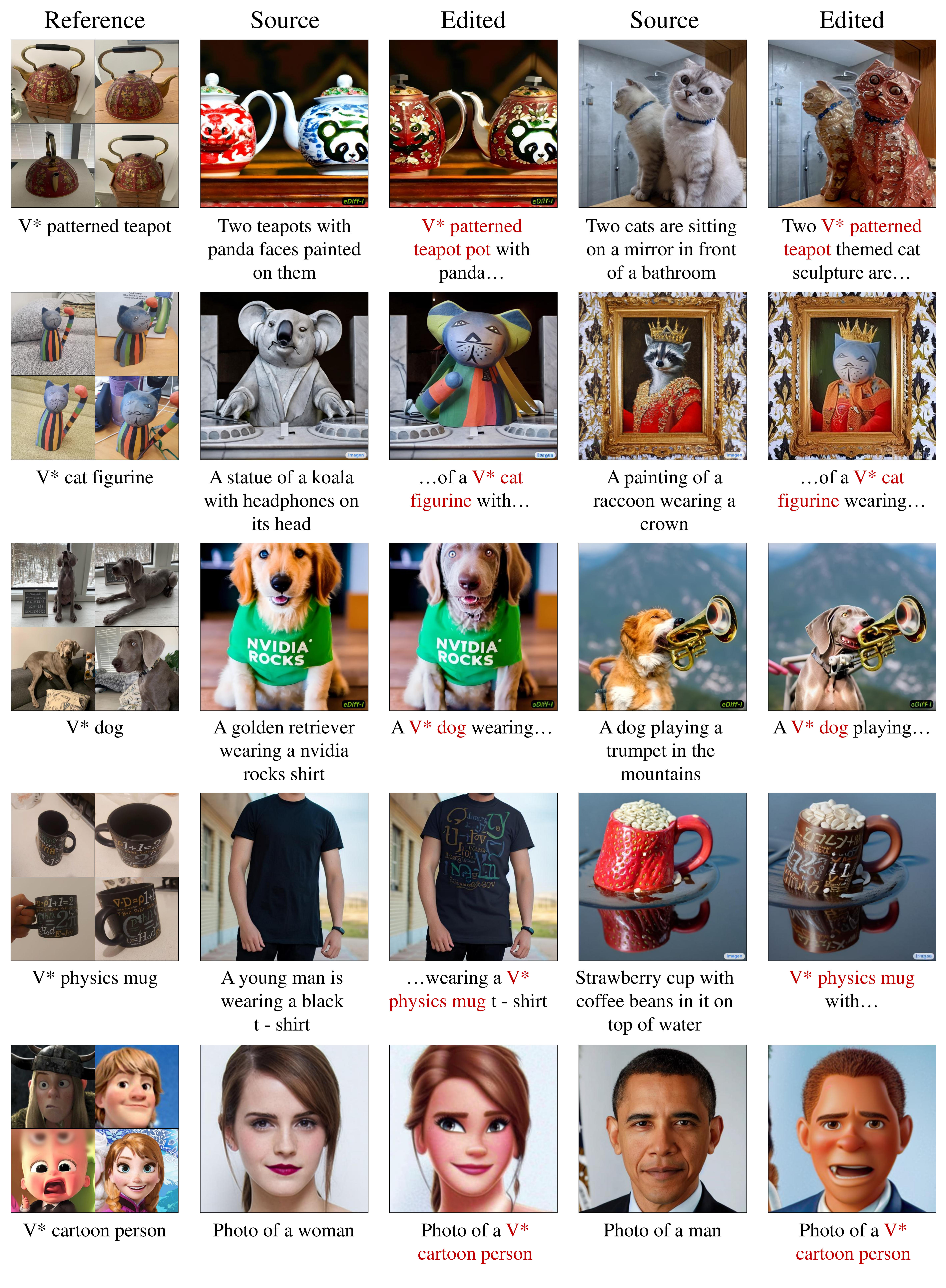}
\end{center}
   \caption{\textbf{Additional Custom-Edit results.}}
\label{fig:main2}
\end{figure*}

Additional Custom-Edit results are shown in~\cref{fig:main2}.

\subsection{Qualitative Comparisons}
\label{sec:appendix-qualitative}
\begin{figure*}
\begin{center}
%\fbox{\rule{0pt}{2in} \rule{.9\linewidth}{0pt}}
   \includegraphics[width=0.85\linewidth]{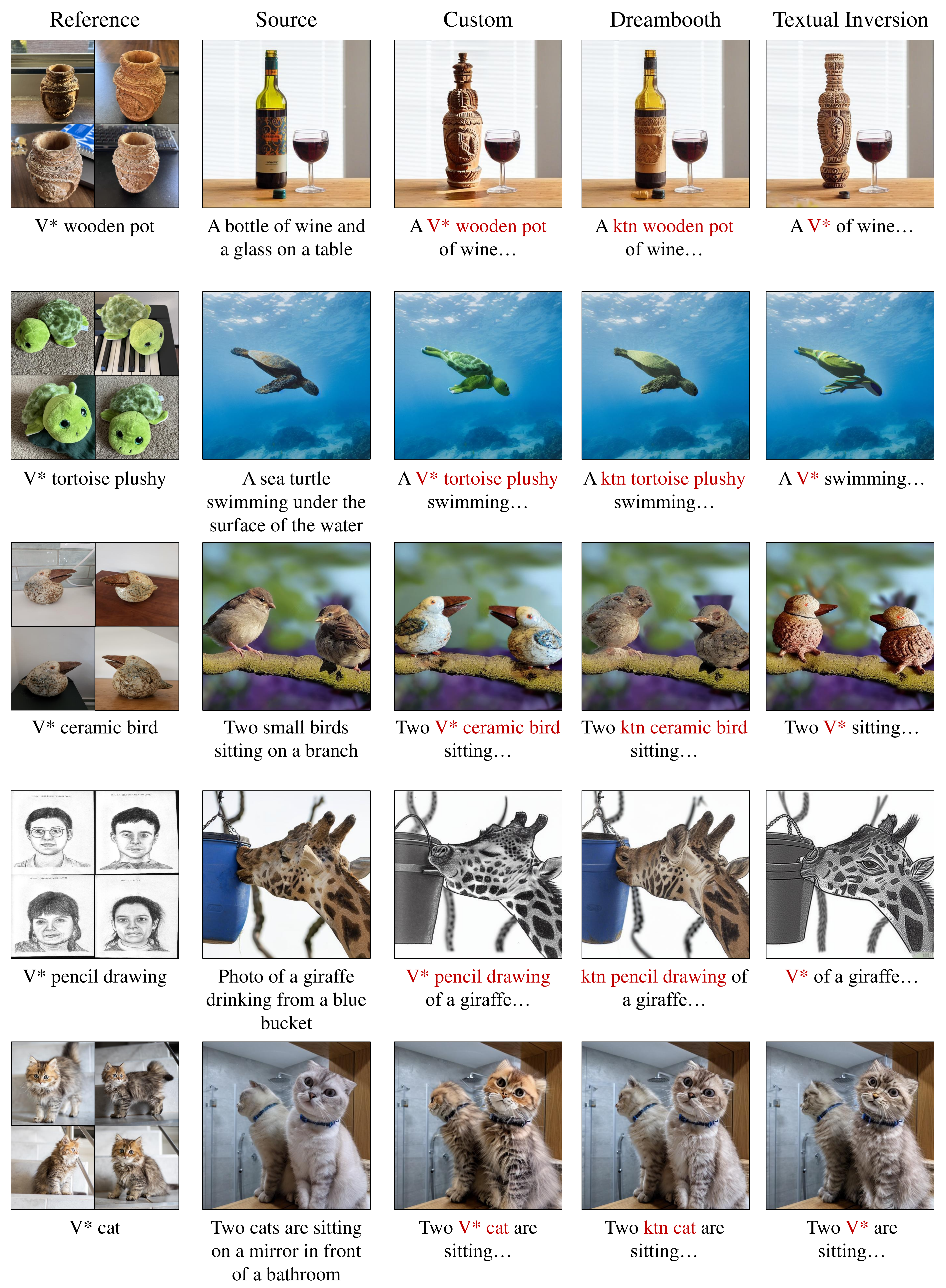}
\end{center}
   \caption{\textbf{Qualitative comparison on P2P.} While the Custom-Diffusion successfully transfers the reference to the source image, both Dreambooth and Textual Inversion fail to capture the precise appearance of the reference. On each row, we use the same strength. We use strength [0.4, 0.8, 0.2, 0.4, 0.8] in order from the first row.}
\label{fig:compare-p2p}
\end{figure*}

\begin{figure*}[t]
\begin{center}
   \includegraphics[width=0.85\linewidth]{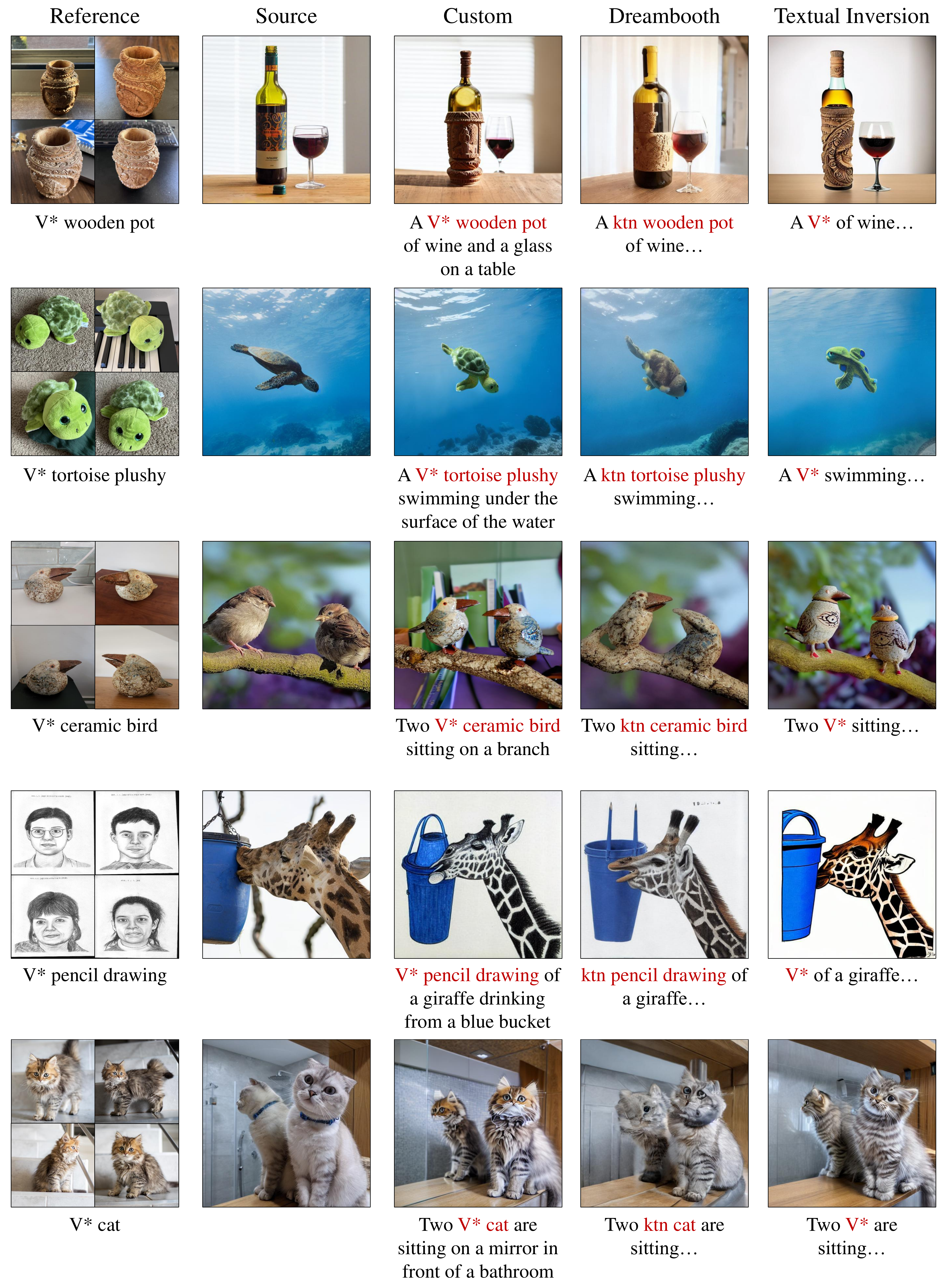}
\end{center}
   \caption{\textbf{Qualitative comparison on SDEdit.} Custom-Diffusion successfully edits the source with some changes in the background since SDEdit diffuses the background as well. Dreambooth and Textual Inversion fail to capture the precise appearance of the reference. On each row, we use the same strength. We use strength [0.8, 0.6, 0.6, 0.8, 0.5] in order from the first row.}
\label{fig:compare-sdedit}
\end{figure*}

Comparisons of customization methods on P2P are shown in \cref{fig:compare-p2p}. Dreambooth fails to modify the source images. Textual Inversion results do not capture details of the references.
Comparisons on SDEdit are shown in \cref{fig:compare-sdedit}. Similar to the results on P2P, Dreambooth and Textual Inversion fail to capture the detailed appearance of the reference.

\subsection{Strength Control}
\label{sec:appendix-strength}

\begin{figure*}[t]
\begin{center}
   \includegraphics[width=0.8\linewidth]{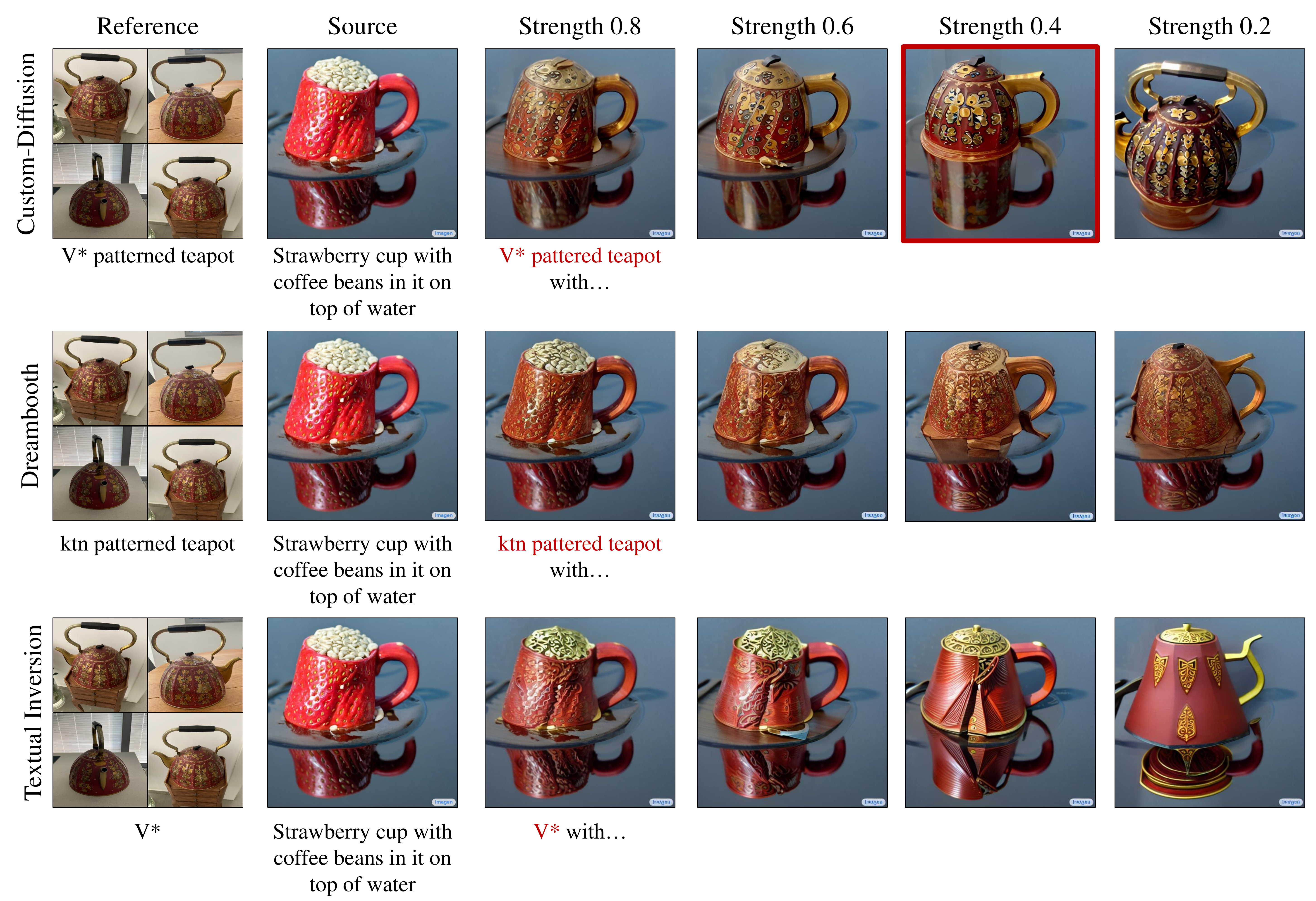}
\end{center}
   \caption{\textbf{Varying strength of P2P.} Red box indicates the best sample.}
\label{fig:strength-p2p}
\end{figure*}

\begin{figure*}[t]
\begin{center}
   \includegraphics[width=0.8\linewidth]{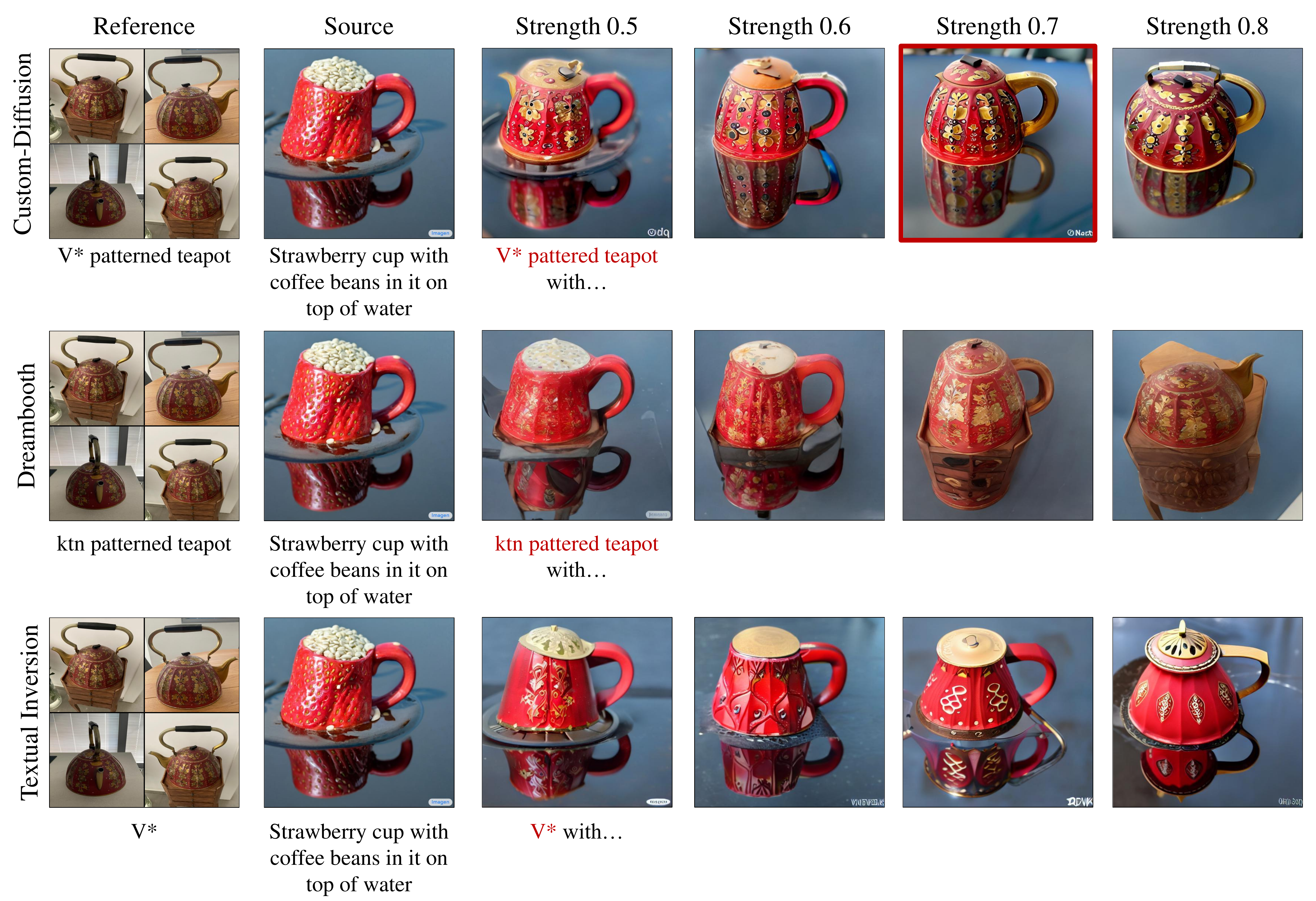}
\end{center}
   \caption{\textbf{Varying strength of SDEdit.} Red box indicates the best sample.}
\label{fig:strength-sdedit}
\end{figure*}

We show how the strength of P2P (\cref{fig:strength-p2p}) and SDEdit (\cref{fig:strength-sdedit}) affect the results. By controlling the strength of these methods, users can choose samples that match their preferences. Our empirical findings suggest that P2P strengths between 0.4 and 0.6, and SDEdit strengths between 0.6 and 0.7 produce good samples. 

\subsection{Failure Cases}
\label{sec:appendix-failure}

\begin{figure}[t]
\begin{center}
   \includegraphics[width=1.0\linewidth]{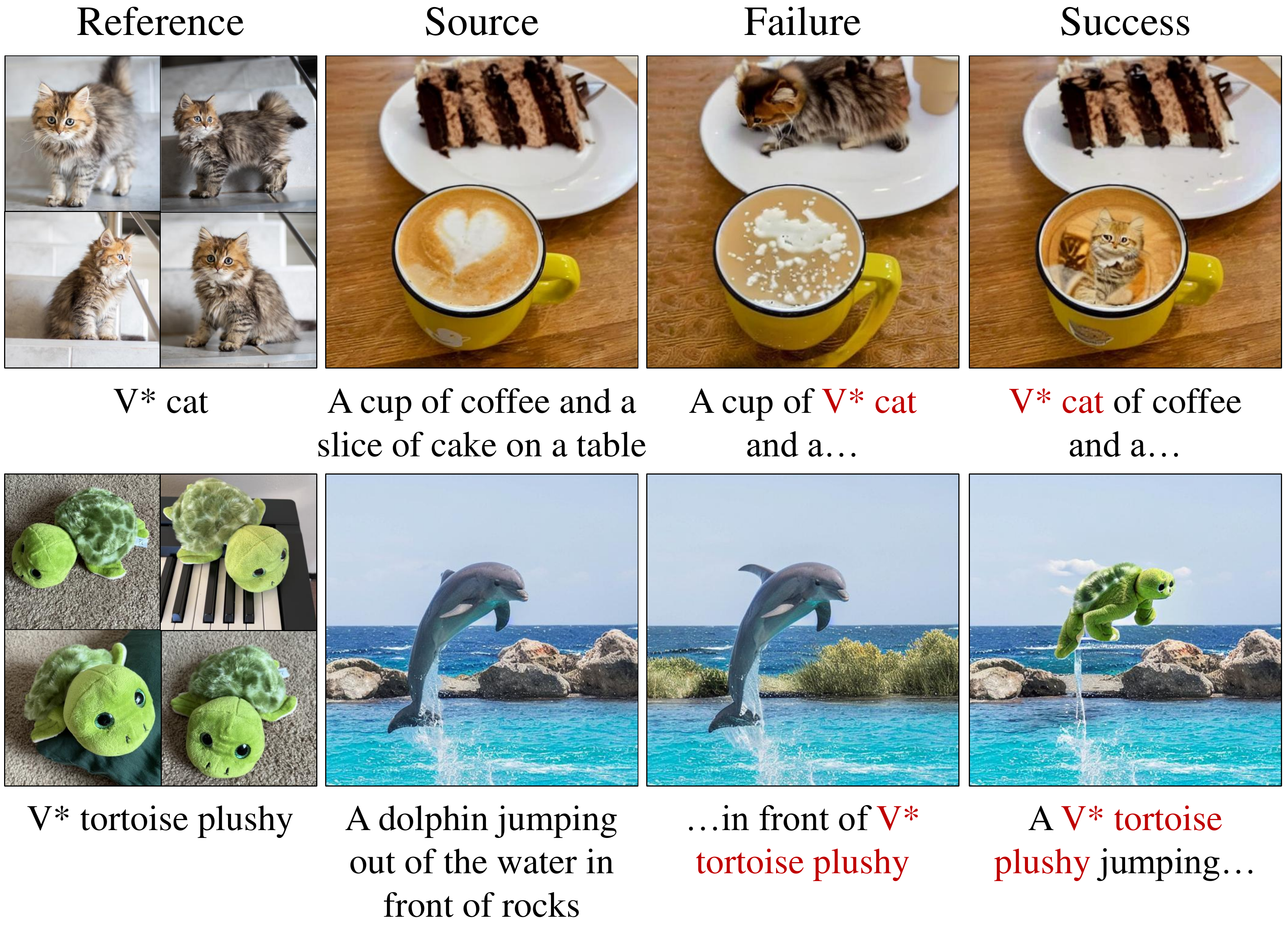}
\end{center}
   \caption{\textbf{Failure cases.}}
\label{fig:failure}
\end{figure}

Failure cases are shown in~\cref{fig:failure}. In the first row, the cake turns into the V* cat instead of the coffee. Replacing `A cup' with `V* cat' resolves the issue. We speculate that Stable Diffusion is not familiar with a cat sitting in coffee, which causes the word 'cat' to fail to attend to coffee. Recent works~\cite{hertz2022prompt,mokady2022null} have noted that attention maps of Stable Diffusion are less accurate than those of Imagen~\cite{sahariaphotorealistic}.

Turning the dolphin into the V* tortoise plushy in the second row is easy. However, we cannot turn rocks into the V* tortoise plushy. The rocks are scattered in the complex scene so, the model requires clarification on which rock to modify. Applying Custom-Edit on extended text-to-image models such as GLIGEN~\cite{li2023gligen}, which is a model extended to the grounding inputs, may solve this problem.

\subsection{Text Similarity}
\label{sec:appendix-textsim}

\begin{figure}[t]
\begin{center}
   \includegraphics[width=0.8\linewidth]{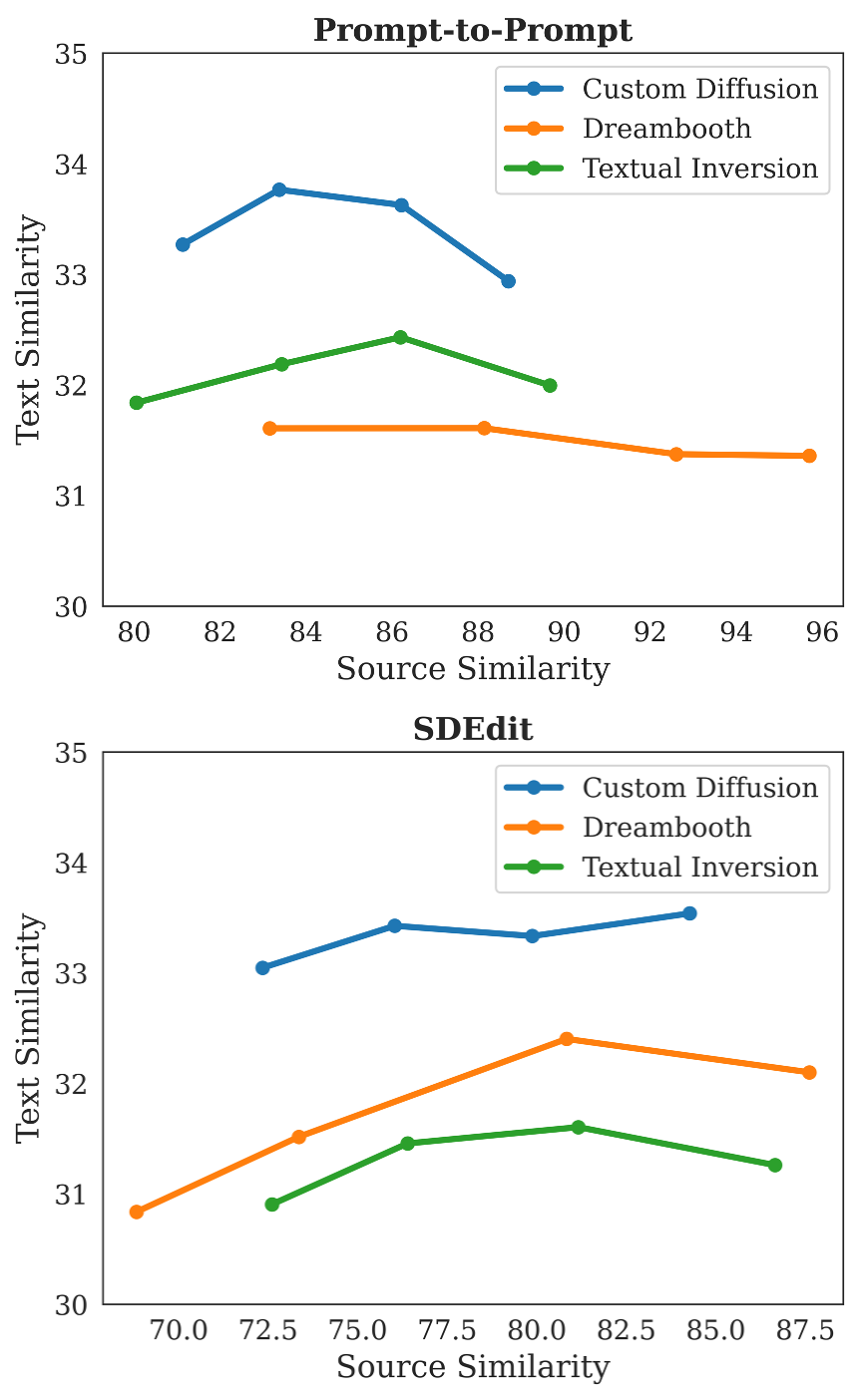}
\end{center}
   \caption{\textbf{Source-Text Similarity Trade-Off.}}
\label{fig:text-tradeoff}
\end{figure}

In addition to the source-reference trade-off shown in the main paper, we show the trade-off between text similarity and source similarity in \cref{fig:text-tradeoff}. We measure text similarity using CLIP ViT-B/32 between the edited image and the target text (with V* omitted). Our improved Custom-Diffusion achieves significantly better text similarity compared to other methods.

\section{Implementation Details}
\label{sec:appendix-details}

\subsection{Customization}
\textbf{Dreambooth and Custom-Diffusion.}
We train a model for 500 optimization steps on a batch size of 2. We use same dataset for prior preservation loss.
During training, we augment text input with the following templates:

\begin{itemize}
\item ``photo of a V* [modifier] [class]''
\item ``rendering of a V* [modifier] [class]''
\item ``illustration of a V* [modifier] [class]''
\item ``depiction of a V* [modifier] [class]''
\item ``rendition of a V* [modifier] [class]''
\end{itemize}

For 1 out of 3 training iterations, we randomly crop images and augment text input with the following templates:

\begin{itemize}
\item ``zoomed in photo of a V* [modifier] [class]''
\item ``close up in photo of a V* [modifier] [class]''
\item ``cropped in photo of a V* [modifier] [class]''
\end{itemize}

We would like to note that for two pet categories (cat and dog), customizing without `[modifier]' token already offered good results.

\textbf{Textual Inversion.}
We train a single token for 2000 optimization steps on a batch size of 4. We used the text template from~\cite{gal2022image}. 

\subsection{Dataset}
\textbf{Reference sets} are collected from prior customization works. Wooden pot, tortoise plushy, cat, and dog from \cite{kumari2022multi}, ceramic bird, cat figurine, patterned teapot from \cite{gal2022image}, and pencil drawing from \cite{ojha2021few}. 

\textbf{Source images} are collected from \href{https://imagen.research.google/}{Imagen}~\cite{sahariaphotorealistic}, \href{https://research.nvidia.com/labs/dir/eDiff-I/}{eDiff-I}~\cite{balaji2022ediffi}, \href{https://imagic-editing.github.io/#}{Imagic}~\cite{kawar2022imagic}, \href{https://muse-model.github.io/}{Muse}~\cite{chang2023muse}, Null-Text Inversion~\cite{mokady2022null}, and Text2Live~\cite{bar2022text2live}.
We provide source-reference pairs used for quantitative comparisons in the \href{https://docs.google.com/spreadsheets/d/1RavwJLn0wiGDa8FIOmFyobey3E4r6xm1/edit?usp=share_link&ouid=101855495888904790577&rtpof=true&sd=true}{supplementary material}.

\section{Additional Background}
\label{sec:appendix-background}
\subsection{Prompt-to-Prompt}
The attention map editing operation $Edit$ includes two sub-operations, namely \textit{prompt refinement} and \textit{word swap}. \textit{word swap} refers to replacing cross-attention maps of words in the original prompt with other words, while \textit{prompt refinement} refers to adding cross-attention maps of new words to the prompt while preserving attention maps of the common words.

\subsection{Null-Text Inversion}
Editing a real image requires deterministic mapping of the image to noise. However, deterministic DDIM inversion~\cite{song2020denoising} fails to reconstruct the source image due to the accumulated error caused by classifier-free guidance~\cite{ho2022classifier}. Null-text inversion~\cite{mokady2022null} addresses this issue by optimizing unconditional text embeddings, which take only a minute for a source image. Note that the diffusion model is not trained; therefore, the model maintains its knowledge.